# Design and Development of an Expert System to Help Head of University Departments

Shervan Fekri-Ershad, Hadi Tajalizadeh, Shahram Jafari

*Abstract-* One of the basic tasks which is responded for head of each university department, is employing lecturers based on some default factors such as experience, evidences, qualifies and etc. In this respect, to help the heads, some automatic systems have been proposed until now using machine learning methods, decision support systems (DSS) and etc. According to advantages and disadvantages of the previous methods, a full automatic system is designed in this paper using expert systems. The proposed system is included two main steps. In the first one, the human expert's knowledge is designed as decision trees. The second step is included an expert system which is evaluated using extracted rules of these decision trees. Also, to improve the quality of the proposed system, a majority voting algorithm is proposed as post processing step to choose the best lecturer which satisfied more experts' decision trees for each course. The results are shown that the designed system average accuracy is 78.88. Low computational complexity, simplicity to program and are some of other advantages of the proposed system.

## 1. INTRODUCTION

In more well known universities of the world, one of basic tasks which is responded for head of each department, is selecting best lecturer for respected courses. Near all of the heads employ a lecturer for a specific course based on some default factors such as teaching experience, quality to communicate with students, evidences and certificates, resume, consensus, C.V. and many others. The number of courses which is taught in each semester is much, and so on, choosing the best collection of the lecturers is a complex task. Moreover, this task should be done every semester periodically.

For example, suppose three volunteers have requested for teaching artificial intelligence (AI) course in a computer department. The first one, have BS.c in software engineering, Ms.c and PhD in artificial intelligence, but he doesn't have any teaching experience. The second have BS.c and MS.c. in software engineering and PhD in image processing. He has taught electronic courses in computer department for 15 years. And the last volunteer has BS.c in software engineering and MS.c in artificial intelligence. He has taught AI course in computer department for 3 years. He is a PhD student right now. If we consider a feature vector for each volunteer which have 4 dimensions where show their certificates and teaching duration, three different vectors are extracted these call F1, F2 and F3 as follows:

F1=<Software, AI, AI, 0>
F2=<Software, Software, Image processing, 15>
F3=<Software, AI, Non, 3>

According to these feature vectors which are extracted for each volunteer based on default factors, it is a complex problem for head of computer department to choose the best between them. In this problem, there are many challenges for head to choose the best such as follows: What dimension is more important? Is PhD certificate necessary for teaching AI or not? Could a volunteer teach AI, who has experience in teaching electronic courses?

In order to solve this problem, an intelligent system is needed which may be done based on artificial intelligence tools such as machine learning and evolutionary algorithms or computer management systems such as decision support systems (DSS). In this respect, our researches are shown that not appropriated algorithm or system is proposed to help department heads right now.

Expert System is software which has the ability to replicate the thinking and reasoning capacity of humans based on some facts and rules presented to it. The use of expert systems finds its place in diverse sectors like medical diagnosis [1], decision support systems [2], biology [3], educational [4], chemical [5], and etc. For example, the use of expert system in medical diagnosis dates back to the early 70's when MYCIN [6], an expert system for identifying bacteria causing diseases was developed at Stanford University.

In this paper, we designed an automatic expert system as helpmate of university department head to choose best lecturer for each course among of the volunteer respectively. The proposed approach is included two main steps. In the first step, some decision trees are made based on facts which extracted of expert humans knowledge who are really heads of university departments. The designed system is a rule based system, so on, in the second step, some rule set are extracted of each made decision trees and finally to choose the best lecturer for each course, the volunteer is selected who satisfy majority rule sets.

## 2. DESIGNED EXPERT SYSTEM

Our designed expert system for helping department heads is a rule based system which has been developed using Clips as implementation software. Each expert system is included two main parts. These are Knowledge Base (KB) and Inference Engine (IE). In the following subsections, these two parts are described completely.

### 2.1. Knowledge Base (KB)

According to description of our problem, the heads should choose best lecturer for each one of the courses individually. In this respect, some facts should be archived for each course in KB. To done it correctly, first of all, we designed a questionnaire which included basic questions about each course. Next, the questionnaire is proposed to some expert humans who are head of departments in universities to answer. For example, a designed questionnaire for computer engineering department is shown in table1. It is answered just for example.

Where, each row is show the name of computer science courses such as Artificial Intelligence (AI), Data Base (DB), Network Security (NS), Algorithm Designing (AD), Computer Networks (CN) and etc. in table1, each column is show a specific question which should be answer by expert man about the relative course.

It is notice that the answer should be in a default area. For example, the expert answer to question.1 should be one of the fields of computer science in Bs.c. such as Software engineering or Hardware engineering. The questions and related areas for answers are as follows:

*Bs.c* = what should be the field of lecturer's Bs.c?= Software / Hardware

*Ms.c* = what should be the field of lecturer's Ms.c?= Artificial Inteligence / Computer Sturcture / Software / Algorithm Designing

*Phd* = what should be the field of lecturer's Phd?= Artificial Inteligence/ Computer Structure / Software / Algorithm Designing

*Corse Score* = what should be the lecturer score on Corse at least? = Range of 10 to 20

*Taught* = what courses the lecturer should be taught in previous = Listed Courses in first column

After completing the questionnaires by human experts, a decision tree must be designed for each questionnaire. The expert's knowledge designing and archiving are done by this process. In this paper, we notified made decision trees by "$DT_i$", which shows the decision tree that is designed by $i_{th}$ expert. The proposed approach is a rule based system, and DT is a full set of rules. Finally, in order to make KB, completed DT's should be archived as part of KB. It is shown in Eq.(1). Where, N is the number of human experts.

KB = $[|DT_1|, |DT_2|, |DT_3|, … … … |DT_N|]$    (1)

In figure1, a decision tree is made for completed questionnaire that is shown in table1.

### 2.2. Inference Engine (IE)

According to the previous section, KB is made by DT's. Each DT is a full set of rules. In this step, to design the inference engine (IE), that tells head, what course the volunteers can teach?, the rules are extracted from DT's. Extracting rule from decision tree is done based on a simple algorithm [7] which is described by Russel and Norvig. Each leaf shows posterior of the rule and path from root to converge leaf is show the antecedent of the rule. In this respect, after extracting rules, the KB is changed as it shown in Eq. (2).

KB = $[\{R_1^1, R_1^2, …, R_1^M\}, \{R_2^1, R_2^2, …, R_2^M\}, … … \{R_N^1, R_N^2, …, R_N^M\},]$    (2)

Where, $R_i^j$ is show the $i_{th}$ rule that is extracted from $j_{th}$ DT. Each rule has some antecedents, each volunteer may be satisfied some of them, so, the rule which has more satisfied antecedents, is considered as winner and its posterior is considered as selected course.

For example, suppose there is a volunteer for lecturing computer courses. The answers of volunteer to questions are shown by a feature vector F1. The rule set that is extracted of DT is shown by Rule-Set (DT).

F1 = <Hardware, AI, AI, (AI=19, DB=20, DT=14), (NS +CN) >

Rule-Set (DT) = {

**Rule 1:**

**If** Bs.c=software & Ms.c=AI & Phd=AI & Score>18 & Taught = AI+AD

**THEN** Course= AI

**Rule 2:**

**If** Bs.c=software & Ms.c=software & Phd=software & Score>15 & Taught = DB

**THEN** Course= DB

**Rule 3:**

**If** Bs.c=hardware & Ms.c= structure & Phd= structure & Score>15 & Taught = NS+CN

**THEN** Course= NS

}

Now, the number of satisfied antecedents of the first rule is 3, respectively satisfied antecedent's number of second rule is 1 and it's 4 for third rule. The amount of accuracy for each course is notified by DT (Course Name). According to the results, for this example, accuracy amount of each course is as follows:

DT (AI) = 3     DT (DB) = 1     DT (NS) = 2

Finally, in this example, the volunteer can lecture AI better than others. In other mean, the designed system is suggested AI course for volunteer. It is show in Eq. (3).

System Suggestion = {i | Max (DT(i))    for i=1,2,…..,C }  (3)

There is a main point about this algorithm. The expert human may answer to a question by more than one value. For example, the expert may answer: "Software or Hardware" to this question: "What should be Bs.c field of lecturer?" In this manner, two individual rules are extracted which have same posterior. Finally, to determine DT (Course), the maximum satisfied antecedent of each rule which have related course as posterior is considered.

### 2.3. Majority Voting

In previous section, the system is designed based on knowledge of one expert human, but the proposed system may be a multi-expert. It means some experts systems can be run parallel and finally their results are integrated. In this respect, if we want to consider knowledge of more than one expert human together, it is enough to voting between their suggested course. Finally, the course is chosen which is suggested by more experts. The full block diagram of the designed system is shown in Figure2.

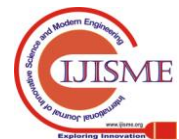

## 3. RESULTS

In this paper a system is designed to help university department heads. In order to evaluate performance of the proposed system, a train dataset is collected using some Iranian university facts. Three different departments which included Computer science, Electronic engineering, Civil Engineering are considered to make train set. In this respect, a train set is made for each department individually. Each train dataset is included 30 instances where each one has 5 dimensions and a label. Each instance is shown a real lecturer who has lectured in university. Each dimension is a fact about him based on the questions. List of questions are same for each department and it is shown in table1 but the answers areas are different. For example, the question – What is your Bs.c is same for all of the departments – but the answer are of these question for computer science is "software and hardware" and it is "Electronic, communication" for electronic engineering department.

The label of each instance is a course name. The universities are guaranteed that each instance (Real lecturer) is lectured the related Course (Label) successfully.

In order to design the proposed system, 5 experts human are answered questionnaire and a specific decision tree is built for each one. After collecting the train sets, the designed system is applied on all of the instances individually. The accuracy of designed system is computed based on Eq.5.

$$Acc = P/Q \quad (4)$$

Where, Q means the number of all instances that is 30, and P means the number of instances that is classified correctly. Correct classification means that the designed system suggestion about them is same with their label. The results are shown in table 2.

As it is shown in table2, the designed system accuracy is maximized for civil engineering department, which means 27 instances are classified correctly between 30. According to the table2, the average accuracy rate of designed system is 78.88.

## 4. CONCLUSION

The main aim of this paper was to design an automatic and efficient system which helps head of university department to employ lecturers and assign better course for them based on their qualities. In this respect, a system is designed using expert system. In order to design the system, decision tree techniques are used which make it robust. Moreover the proposed is rule based which increase flexibility of system. Also, the proposed system is multi-expert which allows it to run some experts systems in a parallel condition and finally, integrated their results to suggest a unique respond. Results have shown the reliability of the proposed approach to suggest best course to volunteers. Low computational complexity, high accuracy, multi-expert, handling numeric and nominal questions together are some of main advantages of our proposed approach.

**Table1.** Some questions of computer science expert to evaluate knowledge base

| Questions / Courses | Bs.c | Ms.c | Phd | Course Score | Taught |
|---|---|---|---|---|---|
| **AI** | Software | Artificial Intelligence | Artificial Intelligence | 18 | AI + AD |
| **DB** | Software | Software | Software | 15 | DB |
| **NS** | Hardware | Computer Structure | Computer Structure | 15 | NS + CN |
| **CN** | Hardware | Computer Structure | Computer Structure | 15 | CN |
| **AD** | Software | Algorithm Designing/Artificial Intelligence | Algorithm Designing | 18 | AD |

**Table2.** Accuracy rate of designed system for three different departments

| Department | Computer Science | Electronic Engineering | Civil engineering |
|---|---|---|---|
| **Accuracy Rate** | 83.33 | 63.33 | 90.00 |

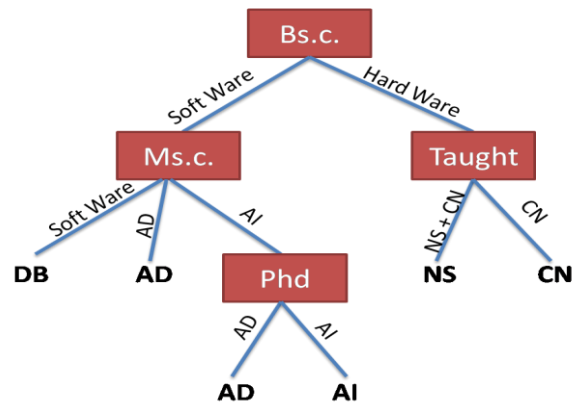

**Figure1.** Decision Tree is made for questionnaire of table1

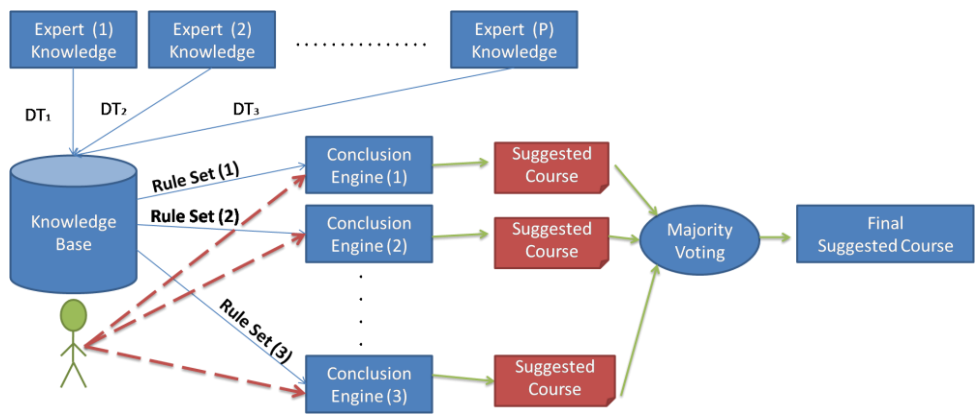

**Figure2**. Diagram of Designed expert system